\documentclass{article}
\usepackage{ijcai18}

\usepackage{times}
\usepackage{xcolor}
\usepackage{soul}
\usepackage[utf8]{inputenc}
\usepackage[small]{caption}
\usepackage{url}  

\usepackage{color}
\usepackage{graphicx}
\usepackage{amsmath}
\usepackage{amssymb}
\usepackage{multirow}

\title{Deep Reasoning with Knowledge Graph for Social Relationship Understanding}

\author{
Zhouxia Wang$^{1,2}$, 
Tianshui Chen$^1$, 
Jimmy Ren$^2$, 
Weihao Yu$^1$,
Hui Cheng$^{1,2}$\thanks{\footnotesize Zhouxia Wang and Tianshui Chen contribute equally to this work and share first-authorship. Corresponding author is Hui Cheng. This work was supported by NSFC-Shenzhen Robotics Projects under Grant U1613211, Key Programs of Guangdong Science and Technology Planning Project under Grant 2017B010116003, and Guangdong Natural Science Foundation under Grant 1614050001452.}
{\normalfont and} Liang Lin$^{1,2}$
\\ 
$^1$ School of Data and Computer Science, Sun Yat-sen University, China \\
$^2$ SenseTime Research, China\\
zhouzi1212,tianshuichen,jimmy.sj.ren,weihaoyu6@gmail.com,\\
chengh9@mail.sysu.edu.cn,
linliang@ieee.org
}

\begin{document}

\maketitle

\begin{abstract}
Social relationships (e.g., friends, couple etc.) form the basis of the social network in our daily life. Automatically interpreting such relationships bears a great potential for the intelligent systems to understand human behavior in depth and to better interact with people at a social level. Human beings interpret the social relationships within a group not only based on the people alone, and the interplay between such social relationships and the contextual information around the people also plays a significant role. However, these additional cues are largely overlooked by the previous studies. We found that the interplay between these two factors can be effectively modeled by a novel structured knowledge graph with proper message propagation and attention. And this structured knowledge can be efficiently integrated into the deep neural network architecture to promote social relationship understanding by an end-to-end trainable Graph Reasoning Model (GRM), in which a propagation mechanism is learned to propagate node message through the graph to explore the interaction between persons of interest and the contextual objects. Meanwhile, a graph attentional mechanism is introduced to explicitly reason about the discriminative objects to promote recognition. Extensive experiments on the public benchmarks demonstrate the superiority of our method over the existing leading competitors.
\end{abstract}

\section{Introduction}
Social relationships are the foundation of the social network in our daily life. Nowadays, as intelligent and autonomous systems become our assistants and co-workers, understanding such relationships among persons in a given scene enables these systems to better blend in and act appropriately. In addition, as we usually communicate via social media like Facebook or Twitter, we leave traces that may reveal social relationships in texts, images, and video \cite{fairclough2003analysing}. By automatically capturing this hidden information, the system would inform users about potential privacy risks. In image analysis tasks, most works are dedicated to recognizing visual attributes \cite{huang2016unsupervised} and visual relationships \cite{lu2016visual}. However, the aforementioned applications require recognizing social attributes and relationships, which receives less attention in the research community. In this work, we aim to address the task of recognizing the social relationships of person pairs in a still image.

\begin{figure}[!t]
   \centering
   \includegraphics[width=0.98\linewidth]{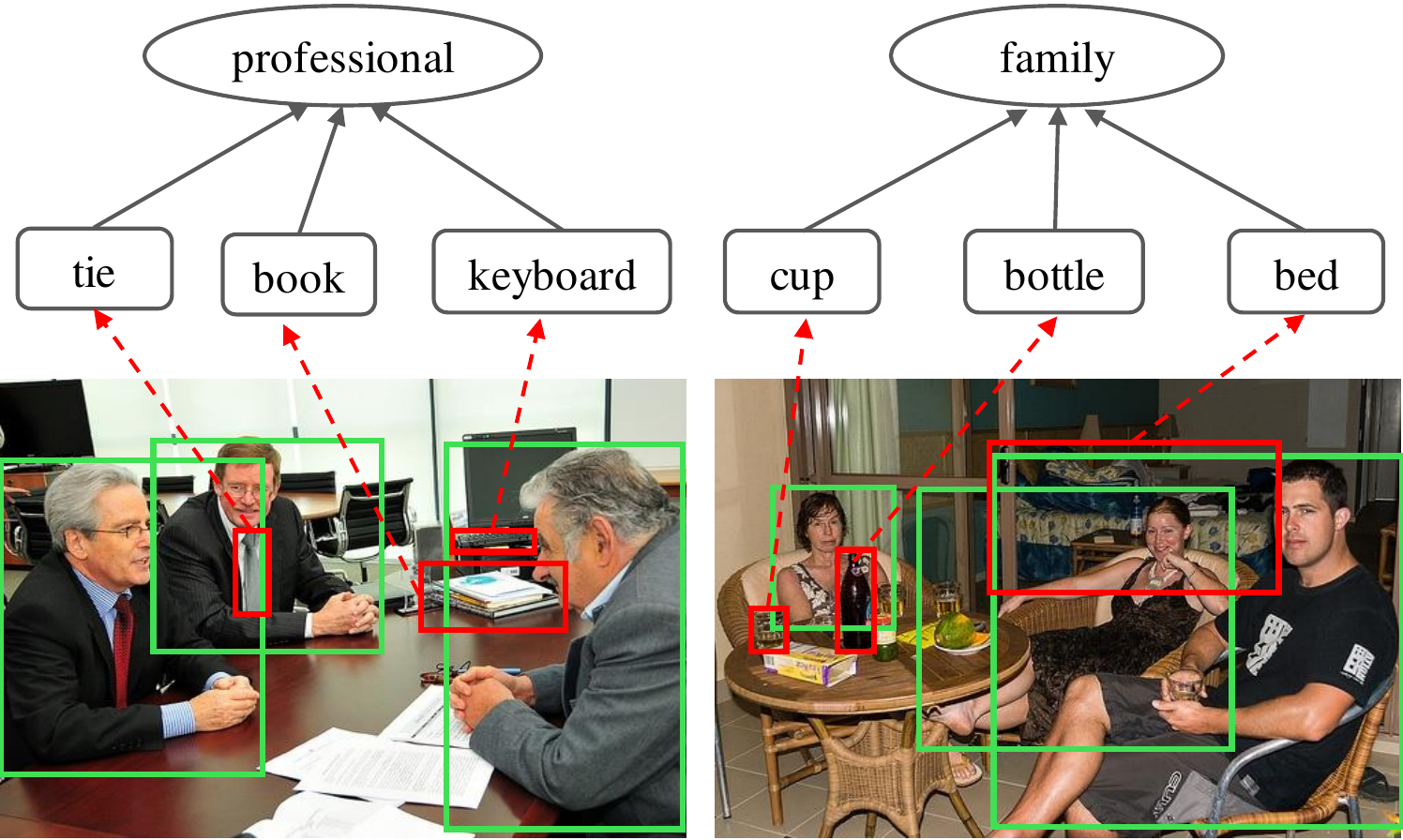} 
   \caption{Two examples of the correlation between social relationship and the contextual objects in the scene.}
   \label{fig:motivation}
\end{figure}

Reasoning about the relationship of two persons from a still image is non-trivial as they may enjoy different relationships in different occasions that contain different contextual cues. For example in Figure \ref{fig:motivation}, given several persons with business wear and with some office supplies such as keyboard around, they are likely to be colleagues. In contrast, if the persons are sitting in a room that has some household supplies like the bed, they tend to be family members. Thus, modeling such correlations between social relationships and contextual cues play a key role in social relationship recognition. Existing works either merely fixate on the regions of persons of interest \cite{sun2017domain} or exploit category-agnostic proposals as contextual information \cite{li2017dual} to perform prediction. Despite acknowledged successes, they ignore the semantic of contextual objects and the prior knowledge of their correlations with the social relationships. Besides, the interaction between the contextual objects and the persons of interest is also oversimplified.

Different from these works, we formulate a Graph Reasoning Model (GRM) that unifies the prior knowledge with deep neural networks for handling the task of social relationship recognition. Specifically, we first organize the prior knowledge as a structured graph that describes the co-occurrences of social relationships and semantic objects in the scene. The GRM then initializes the graph nodes with corresponding semantic regions, and employ a Gated Graph Neural Network (GGNN) \cite{li2015gated} to propagate model message through the graph to learn node-level features and to explore the interaction between persons of interest and the contextual objects. As some contextual objects are key to distinguish different social relationships while some are non-informative or even interferential, we further introduce a graph attention mechanism to adaptively select the most discriminative nodes for recognition by measuring the importance of each node. In this way, the GRM can also provide an interpretable way to improve social relationship recognition by explicitly reasoning about relevant objects that provide key contextual cues. 

In summary, the contributions can be concluded to three-fold. 1) We propose an end-to-end trainable and interpretable Graph Reasoning Model (GRM) that unifies high-level knowledge graph with deep neural networks to facilitate social relationship recognition. To the best of our knowledge, our model is among the first to advance knowledge graph for this task. 2) We introduce a novel graph attention mechanism that explicitly reasons key contextual cues for better social relationship understanding. 3) We conduct extensive experiments on the large-scale People in Social Context (PISC) \cite{zhang2015beyond} and the People in Photo Album Relation (PIPA-Relation) \cite{sun2017domain} datasets and demonstrate the superiority of our methods over the existing state-of-the-art methods. The source codes are available at \url{https://github.com/HCPLab-SYSU/SR}.

\section{Related Work}
We review the related works in term of two research streams: social relationship recognition and graph neural network.
\subsection{Social Relationship Recognition} 

Social relationships form the basic information of social network \cite{li2015celebritynet}. In computer vision community,  social information has been considered as supplementary cues to improve various tasks including multi-target tracking \cite{choi2012unified}, human trajectory prediction \cite{alahi2016social} and group activity analysis \cite{lan2012social,deng2016structure}. For instance, \cite{alahi2016social} implicitly induce social constraint to predict human trajectories that fit social common sense rules. \cite{lan2012social} and \cite{deng2016structure} exploit social roles and individual relation to aid group activity recognition, respectively. 

The aforementioned works implicitly embed social information to aid inference, and there are also some efforts dedicated to directly predicting social roles and relationships. As a pioneering work, \cite{wang2010seeing} use familial social relationships as context to recognize kinship relations between pairs of people. To capture visual patterns exhibited in these relationships, facial appearance, attributes and landmarks are extensively explored for kinship recognition and verification \cite{dibeklioglu2013like,xia2012understanding,chen2012discovering}. To generalize to general social relationship recognition and enable this research, recent works \cite{li2017dual} and \cite{sun2017domain} construct large-scale datasets and employ deep models to directly predict social relationships from raw image input. Concretely, \cite{sun2017domain} build on Bugental’s domain-based theory \cite{bugental2000acquisition} which partitions social life into 5 domains. They derive 16 social relationships based on these domains and extend the People in Photo Album (PIPA) dataset \cite{zhang2015beyond} with 26,915 relationship annotations between person pairs. A simple two-stream model is proposed for social domain/relation recognition. Concurrently, \cite{li2017dual} follow relational models theory \cite{fiske1992four} to define a hierarchical social relationship categories, which involve 3 coarse-level and 6 fine-level relationships. They also build a People in Social Context (PISC) dataset that consists of 22,670 images with 76,568 manually annotated person pairs from 9 types of social relationships, and propose a dual-glance model that exploits category-agnostic proposals as contextual cues to aid recognizing the social relationships.

\begin{figure*}[!t]
   \centering
   \includegraphics[width=1.0\linewidth]{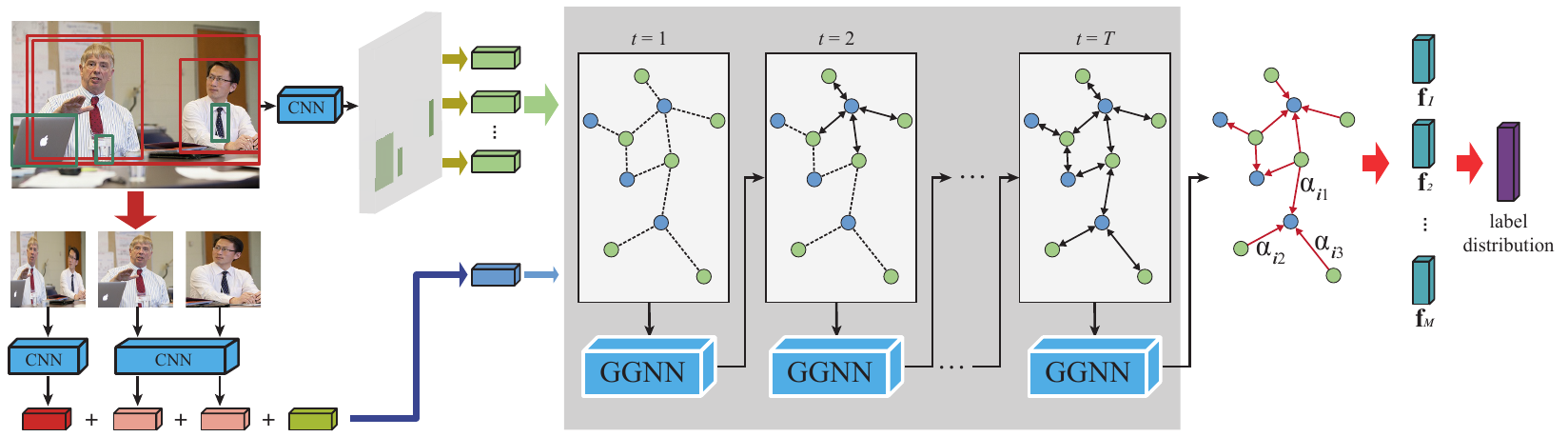} 
   \caption{An overall pipeline of our proposed graph reasoning model. Given an image and a person pair, the GRM initializes the relationship nodes with features extracted from regions of person pair and the object nodes with features extracted from corresponding semantic regions in the image. Then it employs the GGNN to propagate node message through the graph to compute node-level features, and introduces a graph attention mechanism to attend to the most discriminative object nodes for recognition by measuring their importance to each relationship node.}
   \label{fig:GRM}
\end{figure*}

\subsection{Knowledge Representation} 
Representing extr/prior knowledge in the form of graph structure \cite{schlichtkrull2017modeling,lin2017knowledge} and incorporating this structure for visual reasoning \cite{malisiewicz2009beyond,zhu2014reasoning,teney2017graph} has received increasingly attention. For example, Malisiewicz et al. \cite{malisiewicz2009beyond} build a large graph, with the nodes referring to object instances and the edge corresponding to associated types between nodes, to represent and reason about object identities and their contextual relationships. Lin et al. \cite{lin2017knowledge} introduce And-Or graphs for task representation, which can effectively regularize predictable semantic space for effective training. These methods usually involve hand-crafted features and manually-defined rules. 

Recently, more works are dedicated to explore message propagation by learnable neural networks like \cite{chen2018recurrent,wang2017multi} or neural network variants like Graph LSTM \cite{liang2016semantic,liang2017interpretable} and Graph CNN \cite{duvenaud2015convolutional,schlichtkrull2017modeling,kipf2016semi}. \cite{wang2017multi} exploits LSTM network to capture label dependencies by remembering previous information step by step. \cite{liang2016semantic} proposes a Graph LSTM network that propagates message through super-pixels over different level to model their contextual dependencies. Gated Graph Neural Network (GGNN) \cite{li2015celebritynet} is a fully differential recurrent neural network architecture for handling graph-structured data, which iteratively propagate node message through the graph to learn node-level or graph-level representation. Several works have successfully developed GGNN variants for various vision tasks including 3DGNN for RGBD semantic segmentation \cite{qi20173d}, and GSNN for multi-label image recognition \cite{marino2016more}. For example, GSNN learns knowledge representation and concatenates it with image feature to improve multi-label classification. Different from these works, we apply the GGNN to encode the prior knowledge graph and explore the interaction between the person pair of interest and contextual objects to address a newly-raised problem, i.e., social relationship recognition. 

\section{Graph Reasoning Model}
The graph refers to an organization of the correlations between social relationships and semantic objects in the scene, with nodes representing the social relationships and semantic objects, and edges representing the probabilities of their co-occurrences. Given an image and a person pair of interest from the image, the GRM follows \cite{li2017dual} to extract features from the regions of the person pair and initializes the relationship nodes with these features. And it uses a pre-trained Faster-RCNN detector \cite{ren2015faster} to search the semantic objects in the image and extract their features to initialize the corresponding object nodes. Then, the GRM employs the GGNN \cite{li2015gated} to propagate node message through the graph to fully explore the interaction of the persons with the contextual objects, and adopts the graph attention mechanism to adaptively select the most informative nodes to facilitate recognition by measuring the importance of each object node. Figure \ref{fig:GRM} presents an illustration of the GRM.

\subsection{Knowledge Graph Propogation}
GGNN \cite{li2015gated} is an end-to-end trainable network architecture that can learn features for arbitrary graph-structured data by iteratively updating node representation in a recurrent fashion. Formally, the input is a graph represented as $\mathcal{G}=\{\mathbf{V}, \mathbf{A}\}$, in which $\mathbf{V}$ is the node set and $\mathbf{A}$ is the adjacency matrix denoting the connections among these nodes. For each node $v\in \mathbf{V}$, it has a hidden state $\mathbf{h}^t_v$ at timestep $t$, and the hidden state at $t=0$ is initialized by input feature vectors $x_v$ that depends on the task at hand. At each timestep, we update the representation of each node based on its history state and the message sent by its neighbors. Here, we follow the computational process of the GGNN to learn the propagation mechanism.

The graph contains two types of nodes, i.e., social relationship and object nodes, and we initialize their input feature with different contents from the image. Since the regions of the person pair maintain the basic information for recognition, we extract features from these regions to serve as the input features for the social relationship nodes. Similar to \cite{li2017dual}, we first crop three regions, among which one covers the union of the two persons and the other two contain the two persons respectively, and extract three feature vectors from these three regions. These feature vectors, together with the position information encoding the geometry feature of the two persons, are concatenated and fed into a fully connected layer to produce a $d$-dimension feature vector $\mathbf{f}_h \in \mathcal{R}^d$. $\mathbf{f}_h$ is then served as the input features for all the social relationship nodes. 
For the object nodes, we detect the object regions in the image using a pre-trained detector and extract features from these detected regions to initialize the nodes that refer to corresponding categories. As social relationship datasets, e.g., PISC \cite{li2017dual} and PIPA-Relation \cite{sun2017domain}, do not provide object category and their position annotations, the detector cannot be directly trained on these datasets. Fortunately, COCO \cite{lin2014microsoft} is a large-scale dataset for object detection, and it covers 80 common categories of objects that occur frequently in our daily life; thus we get a faster RCNN detector \cite{ren2015faster} trained on this dataset for the collection of semantic objects . And we regard the object with a detected score higher than a pre-defined threshold $\epsilon_1$ as the semantic objects existed in the given image. For the node referring to the object $o$ that is detected in the image, its input feature is initialized by the features extracted from the corresponding region $\mathbf{f}_o \in \mathcal{R}^d$, and otherwise, it is initialized by a $d$-dimension zero vector. 
In addition, we use a one-hot vector to explicitly distinguish the two node types, with $[1, 0]$ and $[0, 1]$ denoting the social relationship and object nodes, respectively, and concatenate them with the corresponding features to initialize the hidden state at timestep $t=0$, expressed as
\begin{equation}
\mathbf{h}^0_v=
\begin{cases}
[[1,0], \mathbf{f}_h] & \text{if $v$ refers to a relationship}\\
[[0,1], \mathbf{f}_o] & \text{if $v$ refers to category $o$ that is detected}\\
[[0,1], \mathbf{0}_d] & \text{otherwise}
\end{cases},
\end{equation}
where $\mathbf{0}_d$ is a zero vector with dimension of $d$. At each timestep, the nodes first aggregate message from its neighbors, expressed as
\begin{equation}
    \mathbf{a}_v^t=\mathbf{A}_v^\top[\mathbf{h}_1^{t-1} \dots \mathbf{h}_{|\mathbf{V}|}^{t-1}]^\top+\mathbf{b},
   \label{eq:init}
\end{equation}
where $\mathbf{A}_v$ is the sub-matrix of $\mathbf{A}$ that denotes the connection of node $v$ with its neighbors. Then, the model incorporates information from the other nodes and from the previous timestep to update each node’s hidden state through a gating mechanism similar to the Gated Recurrent Unit \cite{cho2014learning,li2015gated}, formulated as
\begin{equation}
   \begin{split}
    \mathbf{z}_v^t=&{}\sigma(\mathbf{W}^z{\mathbf{a}_v^t}+\mathbf{U}^z{\mathbf{h}_v^{t-1}}) \\
    \mathbf{r}_v^t=&{}\sigma(\mathbf{W}^r{\mathbf{a}_v^t}+\mathbf{U}^r{\mathbf{h}_v^{t-1}}) \\
    \widetilde{\mathbf{h}_v^t}=&{}\tanh\left(\mathbf{W}{\mathbf{a}_v^t}+\mathbf{U}({\mathbf{r}_v^t}\odot{\mathbf{h}_v^{t-1}})\right) \\
    \mathbf{h}_v^t=&{}(1-{\mathbf{z}_v^t}) \odot{\mathbf{h}_v^{t-1}}+{\mathbf{z}_v^t}\odot{\widetilde{\mathbf{h}_v^t}}
   \end{split}
   \label{eq:ggnn}
\end{equation}
where $\sigma$ and $\tanh$ are the logistic sigmoid and hyperbolic tangent functions, respectively, and $\odot$ denotes the element-wise multiplication operation. In this way, each node can aggregate information from its neighbors while transfer its own message to its neighbors, enabling the interaction among all nodes. An example propagation process is illustrated in Figure \ref{fig:GRM}. After $T$ interactions, the node message has propagated through the graph, and we can get the final hidden state for each node, i.e., $\{\mathbf{h}_1^T, \mathbf{h}_2^T, \dots, \mathbf{h}_{|\mathbf{V}|}^T\}$. Similar to \cite{li2015gated}, we employ an output network that is implemented by a fully-connected layer, to compute node-level feature, expressed by

\begin{equation}
\mathbf{o}_v=o(\left[\mathbf{h}_v^T, \mathbf{x}_v \right]), v=1,2,\dots,|\mathbf{V}|.
\end{equation}

\subsection{Graph Attention Mechanism}
After computing features for each node, we can directly aggregate them for recognition. However, we found that some contextual objects play key roles to distinguish different relationships while some objects are non-informative or even incur interference. For example, the object ``desk'' co-occurs frequently with most social relationships; thus it can hardly provide information for recognition. To address this issue, we introduce a novel graph attention mechanism that adaptively reasons about the most relevant contextual objects according to the graph structure. For each social relationship and neighbor object pair, the mechanism takes their last hidden states as input and computes a score denoting the importance of this object to the relationship. We describe this module formally in the following.

For illustration convenience, we denote the social relationship nodes as $\{r_1, r_2, \dots, r_M\}$ and the object node as $\{o_1, o_2, \dots, o_N\}$, where $M$ and $N$ is the number of the two type nodes, respectively. And their hidden states can be denote as $\{\mathbf{h}_{r_1}^T, \mathbf{h}_{r_2}^T, \dots, \mathbf{h}_{r_M}^T\}$ and $\{\mathbf{h}_{o_1}^T, \mathbf{h}_{o_2}^T, \dots, \mathbf{h}_{o_N}^T\}$ accordingly. Given a relationship $r_i$ and an object $o_j$, we first fuse their hidden states using low-rank bilinear pooling method \cite{kim2016hadamard}
\begin{equation}
\mathbf{h}_{ij}=\tanh(\mathbf{U}^ah_{r_i})\odot\tanh(\mathbf{V}^ah_{o_j}),
\end{equation}
where$\mathbf{U}^a$ and $\mathbf{V}^a$ are the learned parameter matrixes. Then, we can compute the attention coefficient
\begin{equation}
e_{ij}=a(\mathbf{h}_{ij})
\end{equation}
that indicates the importance of object node $o_j$ to relationship node $r_i$. $a$ is the attentional mechanism that is used to estimate the attention coefficient and it is implemented by a fully-connected layer. The model allows to attend on every object nodes but such mechanism ignores the graph structure. In this work, we inject the structure information into the attention mechanism by only computing the attention coefficient $e_{ij}$ for object nodes $j\in \mathcal{N}_i$, where $\mathcal{N}_i$ is the neighbor set of node $i$ in the graph. The coefficients are then normalized to $(0, 1)$ using a sigmoid function
\begin{equation}
\alpha_{ij}=\sigma(e_{ij}).
\end{equation}
$\alpha_{ij}$ is assigned to zero if object node $j$ does not belong to $\mathcal{N}_i$.

Once obtained, we utilize the normalized attention coefficients to weight the output features of the corresponding nodes and aggregate them for final recognition. Specifically, we denote the output features of the social relationship nodes and the object nodes as $\{\mathbf{o}_{r_1}, \mathbf{o}_{r_2}, \dots, \mathbf{o}_{r_M}\}$ and $\{\mathbf{o}_{o_1}, \mathbf{o}_{o_2}, \dots, \mathbf{o}_{o_N}\}$. For relationship $r_i$, we concatenate the features of its own node and the weighted features of context nodes to serve as its final features, that is 
\begin{equation}
\mathbf{f}_{i}=[\mathbf{o}_{r_i}, \alpha_{i1}\mathbf{o}_{o_1}, \alpha_{i_2}\mathbf{o}_{o_2}, \dots, \alpha_{iN}\mathbf{o}_{o_N}].
\end{equation}
Then the feature vector $\mathbf{f}_{i}$ is fed into a simple fully-connected layer to compute a score
\begin{equation}
s_{i}=\mathbf{W}\mathbf{f}_{i}+b
\end{equation}
that indicates how likely the person pair is of social relationship $r_i$. The process is repeated for all the relationship nodes to compute the score vector $\mathbf{s}=\{s_1, s_2, \dots, s_M\}$. 

\subsection{Optimization} 
We employ the cross entropy loss as our objective function. Suppose there are $K$ training samples, and each sample is annotated with a label $y_k$. 
Given the predicted probability vector $\mathbf{p}_k$
 \begin{equation}
      p_{i}^k= \frac{\exp(s_{i}^k)}{\sum_{i'=1}^{M}\exp(s^{k}_{i'})} \ i=1,2,\dots,M,
\end{equation}
the loss function is expressed as
 \begin{equation}
      \mathcal{L}=-\frac{1}{K}\sum_{k=1}^K\sum_{i=1}^{K}\mathbf{1}(y_k=i)\log{p_i^k},
\end{equation}
where $\mathbf{1}(\cdot)$ is the indicator function whose value is 1 when the expression is true, and 0 otherwise. 

\begin{figure}[!t]
   \centering
   \includegraphics[width=1.0\linewidth]{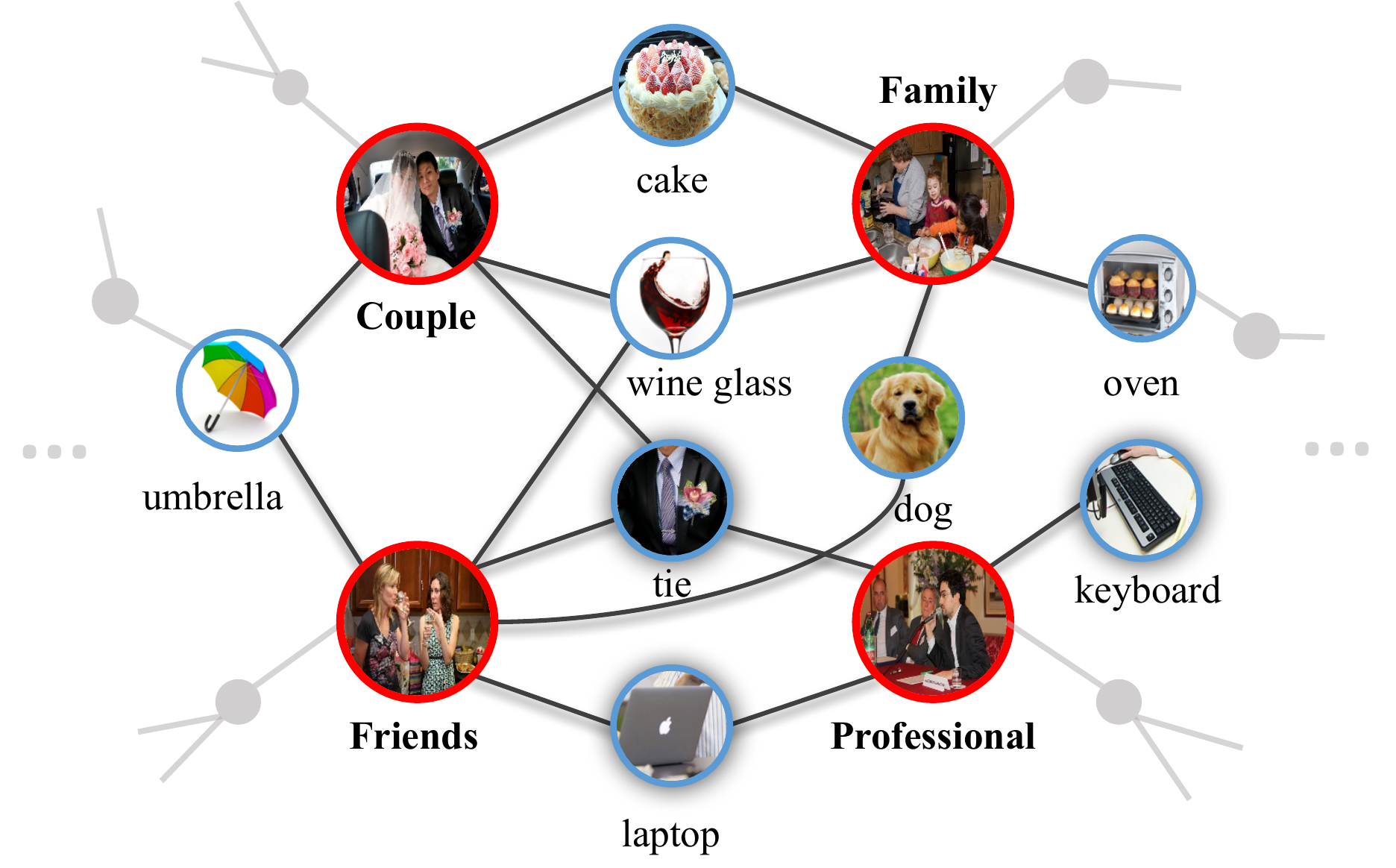} 
   \caption{An example knowledge graph for modeling the co-occurrences between social relationships (indicated by the red circles) and semantic objects (indicated by the blue circles) on the PISC dataset.}
   \label{fig:kg}
\end{figure}

\section{Experiments}
\subsection{Knowledge Graph Building}
The knowledge graph describes the co-occurrences of social relationships and semantic objects in the scene. Building such a graph requires annotations of both social relationships for person pairs and objects existed in the images. However, there is no dataset that meets these conditions. As discussed above, detector trained on the COCO \cite{lin2014microsoft} dataset can well detect semantic objects that occur frequently in our daily life. Thus, we also use the faster RCNN \cite{ren2015faster} detector trained on COCO dataset to detect objects of the image on the social relationship dataset. We regard the detected object with a confidence score higher than a threshold $\epsilon_2$ as the semantic objects in the given image. Here, we utilize a high threshold to avoid incurring too much false positive samples (i.e., $\epsilon_2=0.7$). In this way, we can obtain several pseudo object labels for each image. We then count the frequency of the co-concurrences of each relationship-object pair over the whole training set. All scores are normalized to $[0, 1]$ and the edge with a small normalized score is pruned. Despite the mistakenly predicted labels, the obtained knowledge graph can basically describe the correlation between relationship and object by counting their co-occurrence over a quite large dataset. Besides, by explicitly reasoning about the most important nodes, our GRM can leverage the noise knowledge graph to aid recognition. Figure \ref{fig:kg} illustrate an example knowledge graph for the PISC dataset.

\subsection{Experiment Setting}

\begin{figure}[!t]
   \centering
   \includegraphics[width=0.9\linewidth]{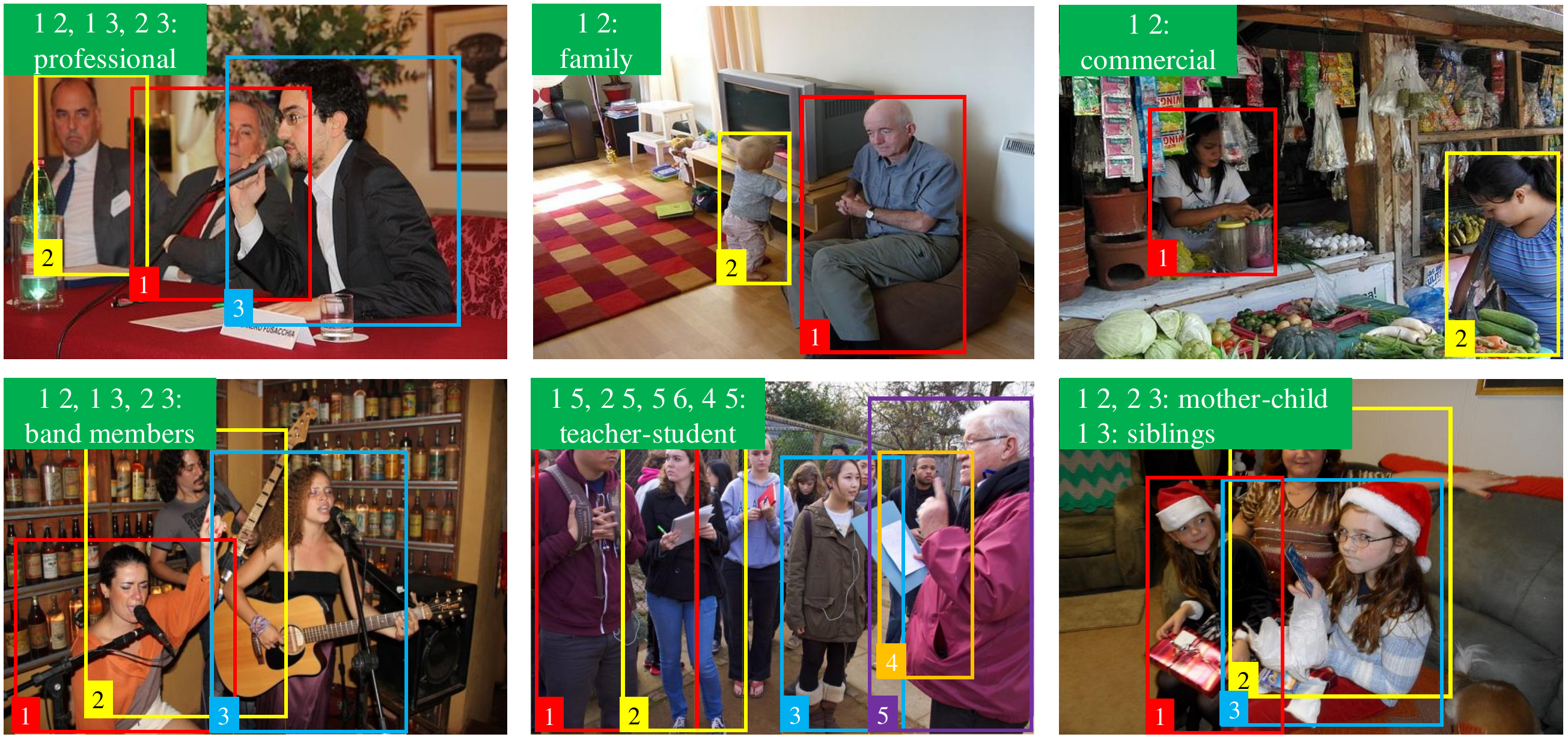} 
   \caption{Samples and their relationship annoations from the PISC (the first line) and PIPA-Relation (the second line) datasets.}
   \label{fig:samples}
\end{figure}

\noindent\textbf{Dataset. }We evaluate the GRM and existing competing methods on the large-scale People in Social Context (PISC) \cite{li2017dual} and the People in Photo Album Relation (PIPA-Relation) \cite{sun2017domain} datasets. The PISC dataset contains 22,670 images and involves two-level relationship recognition tasks: 1) \textbf{Coarse-level relationship} focuses on three categories of relationship, i.e., No Relation, Intimate Relation, and Non-Intimate Relation; 2) \textbf{Fine-level relationship} focuses on six finer categories of relationship, i.e., Friend, Family, Couple, Professional, Commercial and No relation. For fair comparisons, we follow the standard train/val/test split released by \cite{li2017dual} to train and evaluate our GRM. Specifically, for the coarse level relationship, it divides the dataset into a training set of 13,142 images and 49,017 relationship instances, a validation set of 4,000 images and 14,536 instances and a test set of 4,000 images and 15,497 instances. For the fine level relationship, the train/val/test set consist of 16,828 images and 55,400 instances, 500 images and 1,505 instances, 1,250 images and 3,961 instances, respectively. The PIPA-Relation dataset partitions social life into 5 domains and derives 16 social relations based on these domains. Still, we focus on recognizing 16 relationships in the experiment. As suggested in \cite{sun2017domain}, this dataset contains 13,729 person pairs for training, 709 for validation, and 5,106 for test. Serval examples of the samples and their relationship annotations from both two datasets are shown in Figure \ref{fig:samples}.

\noindent\textbf{Implementation details. }For the GGNN propagation model, the dimension of the hidden state is set as 4,098 and that of the output feature is set as 512. The iteration time $T$ is set as 3. During training, all components of the model are trained with SGD except that the GGNN is trained with ADAM following \cite{marino2016more}. Similar to \cite{li2017dual}, we utilize the widely used ResNet-101 \cite{he2016deep} and VGG-16 \cite{simonyan2014very} to extract features for person regions and semantic object regions respectively. 

\begin{table*}[!t]
\centering
\small
\begin{tabular}{c|c|c|c|c||c|c|c|c|c|c|c}
\hline
\centering \multirow{2}{*}{Methods}  & \multicolumn{4}{|c||}{Coarse relationships}  & \multicolumn{7}{c}{Fine relationships} \\
\cline{2-12} & \rotatebox{90}{Intimate} & \rotatebox{90}{Non-Intimate} & \rotatebox{90}{No Relation} & \rotatebox{90}{mAP}  & \rotatebox{90}{Friends} & \rotatebox{90}{Family} & \rotatebox{90}{Couple} & \rotatebox{90}{Professional} & \rotatebox{90}{Commercial} & \rotatebox{90}{No Relation} & \rotatebox{90}{mAP}  \\
\hline
\hline
Union CNN \cite{lu2016visual} & 72.1 & 81.8 & 19.2 & 58.4 & 29.9 & 58.5 & 70.7 & 55.4 & 43.0 & 19.6 & 43.5 \\
Pair CNN \cite{li2017dual} & 70.3 & 80.5 & 38.8 & 65.1 & 30.2 & 59.1 & 69.4 & 57.5 & 41.9 & 34.2 & 48.2 \\
Pair CNN + BBox + Union \cite{li2017dual} & 71.1 & 81.2 & 57.9 & 72.2 & 32.5 & 62.1 & 73.9 & 61.4 & 46.0 & 52.1 & 56.9 \\
Pair CNN + BBox + Global \cite{li2017dual}& 70.5 & 80.0 & 53.7 & 70.5 & 32.2 & 61.7 & 72.6 & 60.8 & 44.3 & 51.0 & 54.6 \\
Dual-glance  \cite{li2017dual} & 73.1 & \textbf{84.2} & 59.6 & 79.7 & 35.4 & \textbf{68.1} & \textbf{76.3} & 70.3 & \textbf{57.6} & 60.9 & 63.2 \\
\hline
Ours & \textbf{81.7} & 73.4 & \textbf{65.5} & \textbf{82.8} & \textbf{59.6} & 64.4 & 58.6 & \textbf{76.6} & 39.5 & \textbf{67.7} & \textbf{68.7} \\
\hline 
\end{tabular}
\caption{Comparisons of our GRM with existing state-of-the-art and baseline methods on the PISC dataset. We present the per-class recall for each relationships and the mAP over all relationships (in \%).}
\label{table:pisc-result}
\end{table*}

\subsection{Comparisons with State-of-the-Art Methods}
We compare our proposed GRM with existing state-of-the-art methods on both PISC and PIPA-Relation datasets.

\subsubsection{Performance on the PISC dataset}
We follow \cite{li2017dual} to compare our GRM with baseline and existing state-of-the-art methods on the PISC dataset. Concretely, the competing methods are listed as follow: 

\noindent\textbf{Union CNN }generalizes the model \cite{lu2016visual}, which predicates general relations, to this task. It feeds the union region of the person pair of interest to a single CNN for recognition.

\noindent\textbf{Pair CNN }\cite{li2017dual} consists of two identical CNNs that share weights to extract features for cropped image patches for the two individuals and concatenate them for recognition.

\noindent\textbf{Pair CNN + BBox + Union} \cite{li2017dual} aggregates features from pair CNN, union CNN and BBox that encode the geometry feature of the two bounding boxes for recognition. We also use these features to describe the person pair of interest and initialize the relationship nodes in the graph.

\noindent\textbf{Pair CNN + BBox + global} \cite{li2017dual} extracts the features of the whole image as contextual information to improve Pair CNN.

\noindent\textbf{Dual-glance} \cite{li2017dual} performs coarse prediction using features of Pair CNN + BBox + Union and exploits surrounding proposals as contextual information to refine the prediction.

We follow \cite{li2017dual} to present the per-class recall and the mean average precision (mAP) to evaluate our GRM and the competing methods on both two tasks. The results are reported in Table \ref{table:pisc-result}. By incorporating contextual cues, both Pair CNN + BBox + Union and Pair CNN + BBox + Global can improve the performance. Dual-glance achieves more notable improvement as it exploits finer-level local contextual cues (object proposals) rather than global context. Different from these methods, our GRM incorporates high-level knowledge graph to reason about the relevant semantic-aware contextual information that can provide a more direct cue to aid social relationship recognition, leading to the performance improvement. Specifically, the GRM achieves an mAP of 82.8\% for the coarse-level recognition and 68.7\% for the fine-level recognition, improving the previous best method by 3.1\% and 5.5\% respectively. It is noteworthy that more notable improvement over other methods for the fine-level recognition is achieved than that for the coarse-level recognition. One possible reason is that recognizing fine-level social relationships is more challenging and thus depends more heavily on prior knowledge. 

It is noteworthy that the GRM uses the Faster RCNN \cite{ren2015faster} pre-trained on the COCO dataset \cite{lin2014microsoft} to detect semantic objects to build the knowledge graph. We also use the same detector to detect semantic objects for initializing the contextual object nodes during both training and test stages. Similarly, work \cite{li2017dual} also uses the Faster RCNN pre-trained on the ImageNet detection data for proposal generation and it involves the ImageNet detection data as extra annotation. Thus, both methods incur extra detection annotations and their comparisons are fair.

\begin{table}[htbp]
\centering
\small
\begin{tabular}{c|c}
\hline
\centering  Methods   & accuracy  \\
\hline
\hline
Two stream CNN \cite{sun2017domain}  & 57.2 \\
Dual-Glance \cite{li2017dual} & 59.6 \\
\hline
Ours  & \textbf{62.3} \\
\hline
\end{tabular}
\caption{Comparison of the accuracy (in \%) of our proposed GRM with existing methods on the PIPA-Relation dataset.}
\label{table:pipa}
\end{table}

\subsubsection{Performance on the PIPA-Relation dataset}
On this dataset, we compare our proposed GRM with two existing methods, i.e., \textbf{Two stream CNN} \cite{sun2017domain} that has reported the results on this dataset and \textbf{Dual-Glance} \cite{li2017dual} that performs best among existing methods on the PISC dataset (see Table \ref{table:pisc-result}). As dual-glance does not present their results on this dataset, we strictly follow \cite{li2017dual} to implement it for evaluation. The results are presented in Table \ref{table:pipa}. Still, our GRM significantly outperforms previous methods. Specifically, it achieves an accuracy of 62.3\%, beating the previous best method 2.7\%. 


\subsection{Ablation Study}

\subsubsection{Significance of knowledge graph}
The core of our proposed GRM is the introduction of the knowledge graph as extra guidance. To better verify its effectivity, we conduct an experiment that randomly initializes the adjacency matrix of the graph and re-train the model in a similar way on the PISC dataset. As shown in Table \ref{table:kg}, the accuracies drop from 82.8\% to 81.4\% on the coarse-level task and from 68.7\% to 63.5\% on the fine-level task. These obvious performance drops clearly suggest incorporating the prior knowledge can significantly improve social relationship recognition.

\begin{table}[htbp]
\centering
\begin{tabular}{c|c|c}
\hline
\centering  Methods  & coarse-level & fine level  \\
\hline
\hline
Random matrix & 81.4 & 63.5 \\
Ours  &  82.8 & 68.7\\
\hline
\end{tabular}
\caption{Comparison of the mAP (in \%) of our GRM that initialize the adjacency matrix by scores counting on the training set and by random scores.}
\label{table:kg}
\end{table}

\subsubsection{Analysis on the graph attention mechanism}

Graph attention mechanism is a crucial module of our GRM that can reason about the most relevant contextual objects. Here, we further implement two baseline methods for comparison to demonstrate its benefit. First, we simply remove this module and directly concatenate features of the relationship nodes and all object nodes for recognition. As shown in Table \ref{table:reasoning}, it suffers from an obvious drop in mAP, especially for the fine-level social relationship recognition that is more challenge. Second, we replace the learnt attention coefficients with randomly-selected scores and retrain the model in an identical way. It shows that the performance is even worse than that using features of all nodes as it may attend to nodes that refer to non-informative or interferential objects. 


\begin{table}[htbp]
\centering
\begin{tabular}{c|c|c}
\hline
\centering  Methods  & coarse-level & fine level  \\
\hline
\hline
Random score & 82.0 & 66.8 \\
Ours w/o attention & 82.6 & 67.8  \\
Ours  &  82.8 & 68.7\\
\hline
\end{tabular}
\caption{Comparison of the mAP (in \%) of ours without attention, ours with random score and our full model.}
\label{table:reasoning}
\end{table}

\subsubsection{Analysis on the semantic object detector}
The detector is utilized to search the semantic objects with confidence scores higher than threshold $\epsilon_1$. Obviously, a small threshold may incur false detected objects while a high threshold may lead to contextual cue missing. Here, we conduct experiments with different threshold values for selecting an optimal threshold. As shown in Table \ref{table:detector}, we find that a relative threshold (i.e., 0.3) lead to best results despite disturbance incurred by the false detected objects. One possible reason is that setting $\epsilon_1$ as 0.3 can recall most contextual objects and the attentional mechanism can well suppress this disturbance, thus leading to better performance.

\begin{table}[htp]
\centering
\begin{tabular}{c|c|c|c|c}
\hline
\centering  $\epsilon_1$  & 0.1 & 0.3 & 0.5 & 0.7  \\
\hline
\hline
mAP & 67.8 & 68.7 & 67.2 & 67.3 \\
\hline
\end{tabular}
\caption{Comparison of the mAP (in \%) of our GRM model using different .}
\label{table:detector}
\end{table}

\subsection{Qualitative Evaluation}
\begin{figure}[!t]
   \centering
   \includegraphics[width=1.0\linewidth]{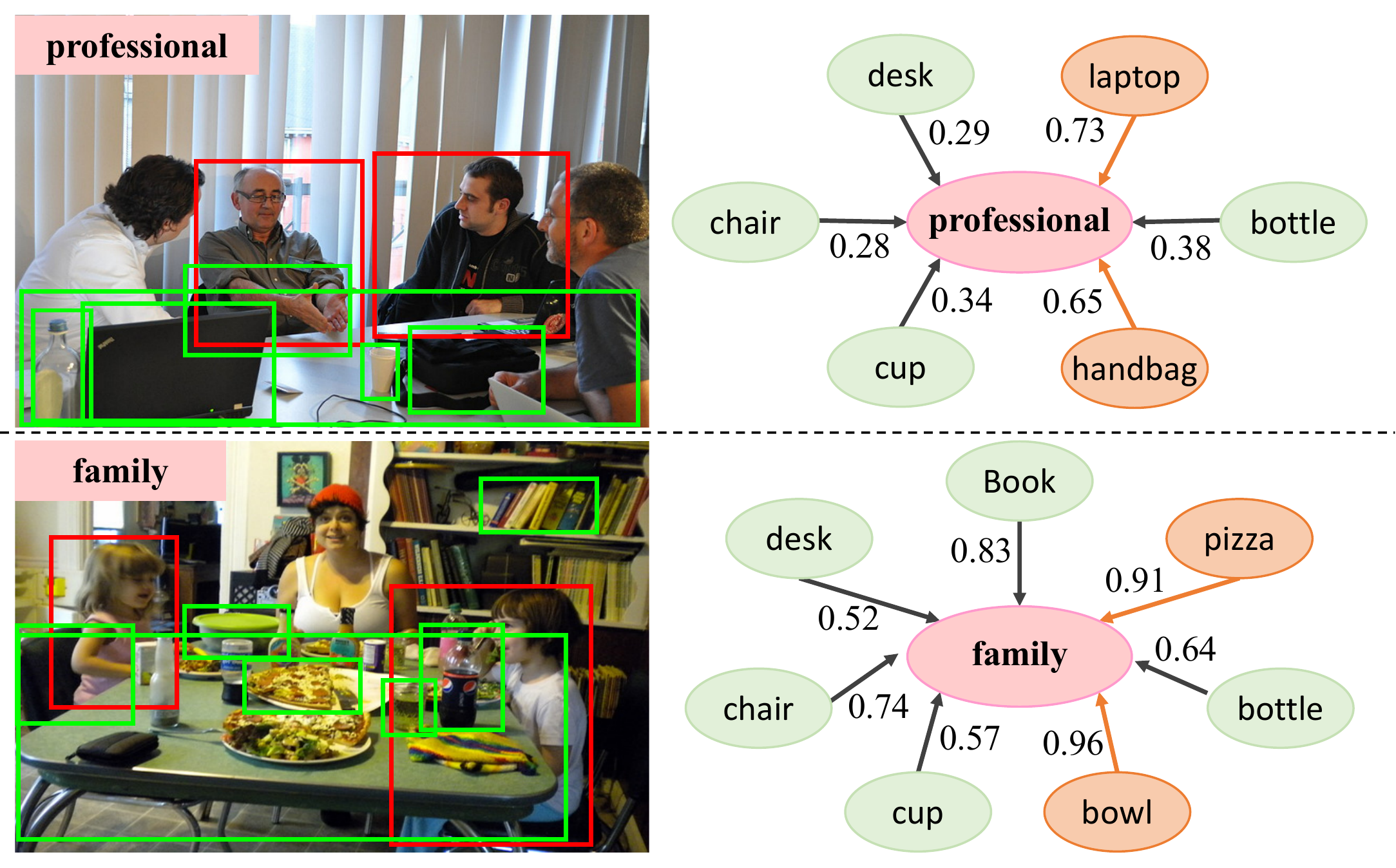} 
   \caption{Two examples of how our GRM recognize social relationships. We visualize the original image, the regions of two persons (indicated by the red boxes), the regions of detected objects (indicated by green boxes) and the ground truth relationships in the left and predicted relationship and nodes referring to the detected objects in the right. The object nodes with top-2 highest scores are highlighted in orange. Best view in color.}
   \label{fig:visualization}
\end{figure}

In this subsection, we present some examples to illustrate how our GRM recognizes social relationships in Figure \ref{fig:visualization}. Considering the first example, the semantic objects including desk, laptop, cup etc., are first detected and utilized to initialize the corresponding object nodes in the graph. However, the objects like desk and cup, co-occur frequently with most social relationships, and thus they can hardly provide informative message to distinguish these relationships. In contrast, office supplies like laptop provide strong evidence for the ``professional'' relationship. Thus, the attention mechanism assigns higher scores to these two nodes, and perform prediction successfully. For the second example, our GRM attends to the bowl and pizza that are the key cues for recognizing ``friend'' relationship.

\section{Conclusion}
In this work, we propose a Graph Reasoning Model (GRM) that incorporates common sense knowledge of the correlation between social relationship and semantic contextual cues in the scene into the deep neural network to address the take of social relationship recognition. Specifically, the GRM consists of a propagation model that propagates node message through the graph to explore the interaction between the person pair of interest and contextual objects, and a graph attention module that measures the importance of each node to adaptively select the most discriminative objects to aid recognition. Extensive experiments on two large-scale benchmarks (i.e., PISC and PIPA-Relation) demonstrate the superiority of the proposed GRM over existing state-of-the-art methods.

\small
\bibliographystyle{named}
\bibliography{ijcai18}

\end{document}